\documentclass[journal]{IEEEtran}

\usepackage{cite}      

\usepackage{graphicx}  

\usepackage{amsmath}   
\usepackage{array}
\newtheorem{example}{Example}

\usepackage{epstopdf}
\usepackage{subfigure}
    
\begin{document}
%
\title{Use of Rapid Probabilistic Argumentation for Ranking on Large Complex Networks}
%
%
\author{Burak~Cetin,
        ~Haluk~Bingol
\thanks{Authors are with the Complex Systems Research Lab, Department of Computer Engineering, Bogazici University, Turkey}
\thanks{This work was partially supported by Bogazici University Research Projects 
under the grant number 07A105}
}

\maketitle

\begin{abstract}
We introduce a family of novel ranking algorithms called \textit{ERank} which run in linear/near linear time and build on explicitly modeling a network as uncertain evidence. The model uses Probabilistic Argumentation Systems (PAS) which are a combination of probability theory and propositional logic, and also a special case of Dempster-Shafer Theory of Evidence. ERank rapidly generates approximate results for the NP-complete problem involved enabling the use of the technique in large networks. We use a previously introduced PAS model for citation networks generalizing it for all networks. We propose a statistical test to be used for comparing the performances of different ranking algorithms based on a clustering validity test. Our experimentation using this test on a real-world network shows ERank to have the best performance in comparison to well-known algorithms including PageRank, closeness, and betweenness.
\end{abstract}

\section{Ranking in Complex Networks}

Ranking nodes in complex networks is an important challenge. Depending on the type of network and the application the meaning of a rank can be different. For the World Wide Web one is usually after popular and informative pages (e.g. Google). For a citation network it is influential papers, for social networks (e.g. Facebook, LinkedIn) it is central/important persons. More recently, networks are tools for calculating trust and transitional trust~\cite{ArtzGil2007}.

Algorithms applied today to large networks often rely on an intuitive idea (e.g. closeness or betweenness centrality ~\cite{FRE79}) or empirical results (e.g. eigenvector based algorithms such as PageRank \cite{PAG98}) but there is no clear and formal foundation as to why they actually work or how they are sound.

When examining a network there is the implicit assumption that it encodes (some uncertain) evidence about the nature of the relations between the nodes. 
Quantitative reasoning under uncertainty is a prolific research field offering many methods and frameworks.

Therefore one expects application of quantitative reasoning to the ranking problem, yet these are rarely used.
There are different reasons for this. 
For example Bayesian networks~\cite{PEA88} are restricted to directed acyclic graphs. 
An alternative is \textit{Dempster-Shafer Theory of Evidence (DST)}\cite{SHA76,DEM68,SHA90} which enjoyed a recent surge of interest~\cite{LIU2003}. 
The adoption of DST based methods have been hampered because of the NP-complete complexity of the computations involved~\cite{HAELEH2003}.
When one contemplates the application of a ranking method to large complex networks such as above, anything much higher than linear time complexity can become virtually impossible to apply. 

In this work we bring forward a family of novel algorithms which we refer to as \textit{ERank}. Our algorithms have linear and lower polynomial time complexities for quantitative reasoning specializing for the node ranking domain. ERank is based on \textit{Probabilistic Argumentation Systems (PAS)}~\cite{HKL00,HAELEH2003_PAS} which are a way of combining propositional logic and probability theory. PAS can be mapped to the DST domain acting as a probabilistic way to interpret DST. 

Our effort can be viewed to have two phases; the construction of a PAS instance to represent a network and the approximation of calculations on that PAS instance. For the first phase, we will use a framework developed by Picard in \cite{PIC00,PIS03,PIC98} and rebrand it as a general PAS based network analysis tool, formalizing our approach in \cite{CET05}. The end product of this phase is a PAS instance. It is a representation of a network in a quantitative reasoning system where one can perform ranking calculations. However as we will explore below, it turns out that it is practically impossible to do the exact PAS calculations required for ranking when a large network is examined due to the NP-complete complexity involved. Essentially, what is needed is a linear or near linear time algorithm when one considers such a task.

In the second phase we introduce ERank as a means of approximating these complex calculations. ERank is a specialized approximation algorithm which works for the PAS instance mapped from a network such as above. It is an iterative algorithm building on the idea of propagating probabilities on the network and rapidly generating estimate results in linear/near linear time.

We view to be an important part of the contribution of this article to be bridging the research in two different fields; ranking algorithms for very large networks and quantitative reasoning. We have strived to keep our text accessible to researchers from both directions.

The remainder of this article will be organized as follows: 
In Section \ref{sec:Background} we will brief well-known and widely used ranking algorithms, present an overview of PAS limiting our focus to directly relevant parts. We will also introduce the Reuters news co-occurrence network~\cite{OZG08} which will be our real world test bed throughout the article. 
Section \ref{sec:ModelWithPAS} will show how a network is mapped to a PAS instance. 
Section \ref{sec:ApproximatingPAS} will introduce and examine different aspects of the ERank algorithms. In Section \ref{sec:ReutersPerformanceEval} we will propose a method for comparing the performances of ranking algorithms on the Reuters network. We will then make a study of various well-known ranking algorithms comparing them to ERank. Finally before concluding we will have Section \ref{sec:ChoosingERankParams} exploring how different choices for parameters in ERank affect performance.

\section{Background}
\label{sec:Background}

\subsection{``Importance'' of nodes in complex networks}
``Importance'' is a concept that is frequently met when dealing with complex networks but it is not always well-defined what is meant.
Depending on the type of network it may mean popularity, reliability/reputation or authority among others.
In this work we have used a variety of well known ``centrality measures'' 
which are also mentioned as ``ranking algorithms''. 
These give a measure of how important a node in the network is.

Arguably the oldest of its kind, ``citation count'' is traditionally used in scientific literature both to asses the importance of an article and the authority of an author. 
Citation networks were shown to be small-world networks where \textit{citation count} is simply the in-degree of a node in a citation network~\cite{NEW03}.

Two common measures of centrality are offered in complex networks literature; closeness and betweenness~\cite{FRE79}. 
\textit{Closeness} measures the shortest distance from a person to every other person.
Here central nodes are the ones which are closest to all other nodes. 
\textit{Betweenness} examines the extent to which a node is situated between others in a network.
It is a measure of how much damage there would be to the connectivity  if a given node is removed from the network.

The famous ranking algorithm called PageRank~\cite{PAG98} establishes the importance of a web page for the Google search engine. 
Along with HITS~\cite{KLE99}, these two algorithms sparked interest in these kind of algorithms in the information retrieval community. 
PageRank originally builds on the intuition that while citation count is a reasonable attempt towards assessing the importance of a document it would be even better to ``extend'' it to take the citer's importance into account. 
PageRanks are simply stationary probabilities for a ``random surfer'' on a directed graph who follows one random link at a time, and has a constant probability of making a random jump to any node. 

PageRank was conjectured to be a useful way of ranking pages and its success has been demonstrated in the success of Google. 
However judging the authority of a web page for evaluations can be a very difficult and costly task requiring questionnaires and manual evaluation. 
In a work by Borodin et al.~\cite{BOR05} such an evaluation is done for PageRank and some other algorithms and PageRank was found not to perform better than citation count. 

Picard, whose PAS model for citation networks we generalize and use in this article, suggests the use of PAS for popularity ranking instead of PageRank~\cite{PIS03}. 
In this work, ranking using PAS is highlighted as a means of generating personalized ranks for each user.

Recently in the ``semantic web'' concept the need to assess important nodes have surfaced again. In a survey of such works~\cite{ArtzGil2007} we see that the ranking algorithms we mention (especially PageRank) or similar ones are used. 

\subsection{Probabilistic Argumentation Systems}

We will be using \textit{Probabilistic Argumentation Systems (PAS)} \cite{HKL00,HAELEH2003_PAS} to model relations between different nodes in a network. PAS use a combination of probability theory and propositional logic building in turn on Dempster-Shafer Theory of Mathematical Evidence (DST)\cite{SHA76,DEM68,SHA90}. As both PAS and DST are broad research topics on their own, we will only be concerned with the necessary parts. We believe Picard does a fine job of summarizing in \cite{PIC98} from which we will heavily borrow below. 

Despite what one might think, propositional logic is capable of expressing
uncertainty. Propositions are normally used to express statements
such as \char`\"{}it is sunny\char`\"{}. A proposition can then take
a truth value depending on the system modeled. Let us introduce a
new class of propositions called \textbf{assumptions}. We will be
using these to express uncertainty on propositions. Let $v_{1}$ be
a proposition stating; \char`\"{}it will rain tomorrow\char`\"{},
and a corresponding assumption $a_{1}$. Consider the following: 

\[
a_{1}\to v_{1}\]

We read it as; \char`\"{}if assumption $a_{1}$ is true then it will
rain tomorrow\char`\"{}, thus effectively \char`\"{}it may rain tomorrow\char`\"{}.
More complex relations can be expressed as propositional sentences,
see Table \ref{cap:PAStable} for examples.

\begin{table}[hbt]

\caption{Knowledge representation in PAS\label{cap:PAStable}.}

\center

\begin{tabular}{|m{1.7cm}|m{2.7cm}|m{3cm}|}
\hline 
Type of knowledge&
Logical representation&
Natural language representation\tabularnewline
\hline
\hline 
a fact
& $v_{1}$
& {}``$v_{1}$is a fact''
\tabularnewline
\hline 
a simple rule
& $v_{1}\to v_{2}$
& {}``$v_{1}$ implies $v_{2}$''
\tabularnewline
\hline 
an uncertain fact
& $a_{1}\to v_{1}$
& {}``if assumption $a_{1}$ is true, \newline then $v_{1}$is true''
\tabularnewline
\hline 
a simple uncertain rule
& ${\rm a}_{{\rm 1}}\to{\rm (v}_{{\rm 1}}\to{\rm v}_{{\rm 2}}{\rm )}$ equivalently
\newline${\rm a}_{{\rm 1}}\wedge{\rm v}_{{\rm 1}}\to{\rm v}_{{\rm 2}}$
& {}``if assumption $a_{1}$ is true, 
\newline then $v_{1}$ implies $v_{2}$''
\tabularnewline
\hline
\end{tabular}
\end{table}

A \textbf{Propositional Argumentation System} 
is a triple $(P$, $A$, $\xi)$ where $P={\{ v_{1},v_{2},...,v_{n}\}}$
is the set of propositions, $A={\{ a_{1},a_{2},...,a_{m}\}}$ is the set of assumptions,
and $\xi$ the knowledgebase. 
$\xi$ can sometimes be specified as
a set $\xi={\{\xi_{1},\xi_{2},...,\xi_{n}\}}$ representing a disjunction
of propositional clauses.
Note that $A \cap P = \emptyset$.

A \textbf{hypothesis} $h$ is any logical formula of interest for
us, with symbols in $A\cup P$. An \textbf{argument} is a conjunction
of assumptions which is said to be in favor (or against) of $h$ if
with its assignment $h$ becomes true (or false). Then the hypothesis
$h$ is said to be supported (or discarded) by the argument. The \textbf{support}
of $h$ with regard to $\xi$ is equal to the disjunction of all the
arguments supporting $h$, and is denoted $SP(h,\xi)$.

So far we have considered the qualitative aspect, it is also possible
to introduce a quantitative judgment by using probability assignments
for assumptions. 
The quadruple $PAS_{P}=(P,A,\xi,\Pi)$ is called a \textbf{Probabilistic Argumentation System (PAS)}, where $\Pi$ represents the probability assignments for assumptions  
(e.g. $\Pi = \left[ p(a_1) ... p(a_m) \right]^T$  where $p(a_i)$ is the probability of $a_i$ being true). 
The probability distributions of all the assumptions are assumed to be stochastically independent.
Thus the probability of a clause is simply the multiplication of the individual probabilities for the assumptions involved
(e.g. for the case $a_1=true$ and $a_2=false$, $p(a_1 \wedge a_2)=p(a_1)(1-p(a_2))$). 

The quantitative value representing the support for an hypothesis is \textbf{degree of support}; denoted $dsp(h,\xi)$. Simply put, it yields a value $0 \leq dsp(h,\xi) \leq 1$ which gives the posterior probability that the hypothesis is supported by the evidence.

Note that an important feature of this kind of knowledge-base is that the $dsp$ function is non-decreasing with additional evidence.
Note also that when a given knowledgebase entails \emph{no contradictions} the following equation holds~\cite{HKL00}:
\begin{equation}
dsp(h,\xi ) = p\left( {SP(h,\xi )}\right)
\label{eq:consistentdsp}
\end{equation}

The $dsp$ value corresponds to belief in the hypothesis in DST. 
PAS represent a special case of DST, and make it possible to interpret belief probabilistically~\cite{HKL00}. 
Thus $dsp$ corresponds to the posterior probability that the hypothesis is true in the system.

\begin{example} 
Consider the following Propositional Argumentation System; assumptions
$A=\{{a_{1},a_{2},a_{3}}\}$, propositions $P={\{ v_{1},v_{2}\}}$,
and the knowledgebase $\xi = \{ \xi_{1}, \xi_{2}, \xi_{3}\}$ where \\
$\xi_{1}:a_{1}\to v_{1}$ \\
$\xi_{2}:a_{2}\to v_{2}$ \\
$\xi_{3}:v_{2} \to (a_{3}\to v_{1})$.

If our hypothesis is $h=v_{1}$, the support for $h$ is the disjunction
of all the arguments which make $v_{1}$ true. After examining the
rules above we can see that $SP(h,\xi)$ is:

\begin{equation}
SP(h,\xi)=a_{1}\vee\left({a_{2}\wedge a_{3}}\right) 
\label{eq:PASexample}\end{equation}

Using an alternative notation $SP(h,\xi)=\left\{ {a_{1},a_{2}\wedge a_{3}}\right\}$.

Let the probability assignments for the assumptions be; $p(a_{1})=0.6$, $p(a_{2})=0.3$, and $p(a_{3})=0.2$. 
We already know the supporting arguments for the hypothesis $v_{1}$. 
However, we can not simply add the corresponding
probabilities because they have to be made disjoint first:

\begin{eqnarray*}
\label{eq:PASkb}
dsp(v_{1},\xi) & = & p(SP(v_{1},\xi))\\
 & = & p(a_{1}\vee\left({a_{2}\wedge a_{3}}\right))\\
 & = & p(a_{1})+p(\neg a_{1}\wedge a_{2}\wedge a_{3})\\
 & = & p(a_{1})+(1-p(a_{1}))\cdot p(a_{2})\cdot p(a_{3})\\
 & = & 0.6+(1-0.6)\cdot0.3\cdot0.2\\
 & = & 0.624
\end{eqnarray*}

\end{example}

\subsection{Co-occurrence Network of Reuters News}
We will be using the co-occurrence network of Reuters news~\cite{OZG08} as a test network for our algorithms. We will be analyzing the ``importance'' of the persons in this network. It is constructed using the Reuters-21578 corpus which contains 21578 Reuters newswire articles which appeared in 1987, mostly on economics. This is a network with 5249 nodes and 7528 edges, where nodes represent individual people and there is an edge between two persons if they appear in an article together. We chose to use edges as unweighted. These people are often well-known or powerful people of their time in politics or business. It was shown in \cite{OZG08} this network exhibits small-world properties, presented along with a study of different well-known ranking algorithms. We use a converted version of this undirected network to a directed network by using two arcs in both directions in place of an edge. The diameter of the undirected network is 13.

\section{Using PAS to Model Network Relations}
\label{sec:ModelWithPAS}

PAS for network analysis were initially used to model and analyze citation networks\cite{PIC00,PIS03,PIC98}. In these works the main problem is enhancing the performance of information retrieval with regards to relevance. Picard introduces a PAS based framework to model network relationships between documents. We will be using this model only generalizing it as a general network analysis tool. We have formalized our approach in \cite{CET05}. Simply, the model no longer models documents and hyperlinks on documents, but it can be nodes and links of any network. We introduce the concept of a transitive relation to establish the context of the analysis.

For example, if we want to model the spread of a contagious disease, then the links could represent the infection probabilities between individuals and the node assumptions would be the initial probabilities that a given individual in the population is already infected. In this setting, the degree of support for a given node proposition would give the posterior probability that a given person is sick given the relations structure between individuals. 
When analyzing the importance of persons in a social network then our transitive relation could be 
``(if person A is linked to person B then) person B is influenced by person A'', 
for WWW it can be 
``(if page A links to page B then) page B is found important/informative by page A''. 
The mathematical model is not affected as long as the relation is transitive. It is debatable what constitutes a transitive relation especially in a social setting. For example, if a person (A) is influenced by another (B) who in turn is influenced by a third person (C) it is nevertheless possible (A) and (C) do not know each other. We can still consider this a transitive relation for this model, if (C) can indirectly influence (A) by influencing (B). It is possible to see how this would happen if there is absolute trust involved. The PAS model is capable though of handling a lower level or uncertain level relation.

A network is mapped into a PAS instance $PAS_{P}=(P, A, \xi, \Pi)$. 
Each node $i$ has a corresponding proposition $v_i \in P$ and an assumption $a_i \in A$.
The link from node $i$ to node $j$ has the link assumption $l_{ij} \in A$. The assumptions represent the chosen \emph{transitive relation}.
Then the knowledge base $\xi$ consists of the disjunction of the following forms: \\
$a_i  \to v_i$ : for each node $i$  \\
$(v_i  \wedge l_{ij} ) \to v_j$ : whenever there is a link from node $i$ to $j$.

The knowledge-base in this model is made of Horn clauses (i.e. sentences of the type $a \wedge b \wedge c \wedge ... \to z$). Finding out the support $SP(v_i)$ can be identified as an inference (argument finding) problem and is known to have linear complexity \cite{RUN03}. Also it entails no contradictions, so Eq.\ref{eq:consistentdsp} holds.



\begin{example} 
\label{ex:SP}

\begin{figure}[htbp]
\centering
\subfigure[] 
{
	\includegraphics[scale=0.5,keepaspectratio=true]{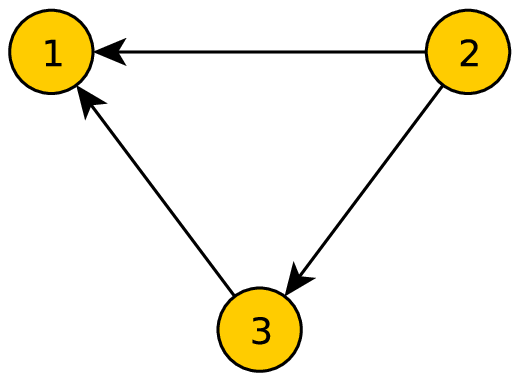}
	\label{fig:ExampleSimplePASgraph1}
}
\hspace{.1in}
\subfigure[] 
{
	\includegraphics[scale=0.5,keepaspectratio=true]{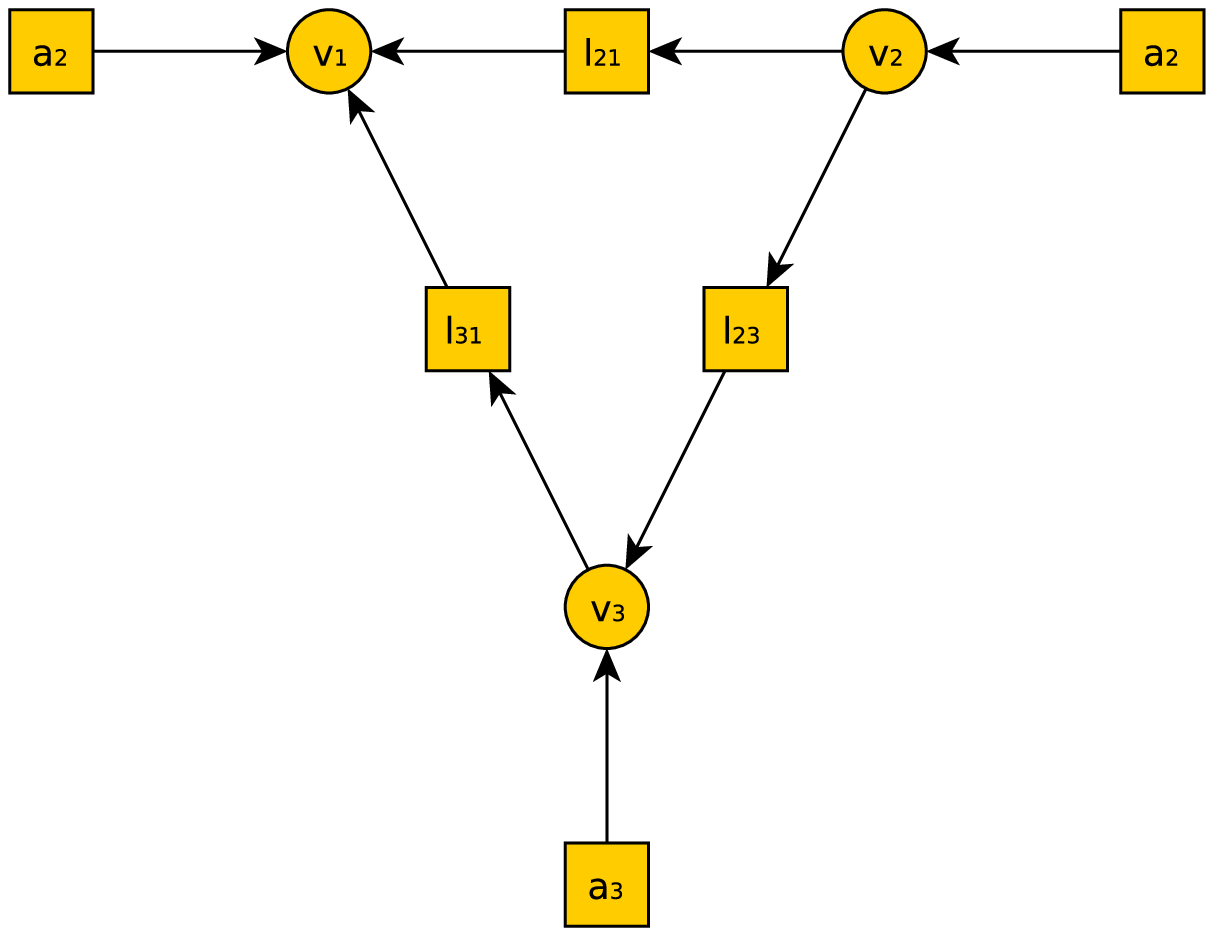}
	\label{fig:ExamplePASgraph1}
}
\caption{
(a) A simple network.
(b) Corresponding PAS graph.
}
\label{fig:Example1} 
\end{figure}

Consider the simple network in Fig.\ref{fig:ExampleSimplePASgraph1}. 
The knowledge-base $\xi$ for this network is given below:\\
$\xi_{1}: a_{1}\to v_{1}$\\
$\xi_{2}: a_{2}\to v_{2}$\\
$\xi_{3}: a_{3}\to v_{3}$\\
$\xi_{4}: (v_2 \wedge l_{21}) \to v_1$\\
$\xi_{5}: (v_2 \wedge l_{23}) \to v_3$\\
$\xi_{6}: (v_3 \wedge l_{31}) \to v_1$\\

Using logical inference on $\xi$ we can find the set of supporting arguments for $v_1$. Note the reach of support of $v_2$ to $v_1$ via $v_3$.
\begin{equation}
\label{eq:PAS_Ex1_PL_SP1}
SP(v_1) = { a_1 \vee (a_2 \wedge l_{21}) \vee (a_2 \wedge l_{23} \wedge l_{31}) \vee (a_3 \wedge l_{31}) }
\end{equation}

\end{example}

Now consider the same network on Fig.\ref{fig:ExamplePASgraph1}, this time also showing the propositional symbols. The circle nodes represent node propositions $v_i$, and the square nodes represent node $a_i$ and link $l_{ij}$ assumptions. Note how the inference process for a given node is reminiscent of walking backwards on the graph from the node.

As proven in \cite{CET05} the general formulation of support for a given node's proposition $v_i$ is:
\begin{equation}
\label{eq:general_sp}
SP(v_i) = a_i  \vee \mathop  \bigvee \limits_{j \in P_i }^{} \left(SP(v_j) \wedge {l_{ji}} \right)
\end{equation}
where $P_i$ is the set containing the parent nodes of $i$.
The \textit{inclusion-exclusion} rule is useful for evaluating this kind of expressions:
\begin{eqnarray*}
p(a \vee b) & = & p(a) + p(b) - p(a \wedge b)\\
\end{eqnarray*}
where $a$ and $b$ are propositional sentences. If $a$ and $b$ are disjunct it becomes:
\begin{eqnarray*}
p(a \vee b) & = & p(a) + p(b) - p(a)p(b)\\
						& = & 1 - (1 - p(a)) (1 - p(b))
\end{eqnarray*}

\begin{example}
\label{ex:PAS_Quantitative}
Now let us look at the quantitative aspect of the previous example. 
We will use the short form $dsp_i$ for $dsp(v_i)$. 
Before we can calculate $dsp_1$, the expression in Eq.\ref{eq:PAS_Ex1_PL_SP1} needs to be made disjoint. Below is one way to do it (dropping $\wedge$s for convenience):

\begin{eqnarray*}
SP(v_1) & = & a_1 \vee a_2 \left( l_{21} \vee l_{23} l_{31} \right) \vee a_3 l_{31}  \\
 & = & a_1 \\
 &   & \vee \neg a_1 a_2 \left( l_{21} \vee l_{23} l_{31} \right)  \\
 &   & \vee \neg a_1 \neg a_2  a_3 l_{31}
\end{eqnarray*}

This sentence is disjoint except the expression in the middle which includes the disjunction of two (disjunct) clauses. 
Using the inclusion-exclusion rule:

$dsp_1  =  p(a_1)$ \\
$  + ( 1 - p(a_1)) p(a_2) \left( {p(l_{21}) + p(l_{23}) p(l_{31}) - p(l_{21}) p(l_{23}) p(l_{31})} \right)$ \\
$  + (1 - p(a_1)) (1 - p(a_2)) p(a_3) p(l_{31})$

Let us use the values $p(a_1)=p(a_2)=p(a_3)=0.3$, and $p(l_{21})=p(l_{31})=p(l_{23})=0.5$. Inserting these above gives $dsp_1 = 0.5047$. 
Using the infection interpretation, when there is a 0.3 probability of ``infection'' on each node, node 1 has a higher posterior probability 0.5047 to eventually catch the disease, which is what we expect to see.
\end{example}

Making an expression disjoint is in fact an NP-complete problem as it involves the satisfiability problem (SAT) which is a well-known NP-complete problem \cite{RAU03}. 
So, although finding $SP(v_i)$ of node $i$ is relatively easy with $O(N)$ complexity, 
finding $dsp_i$ can be prohibitively expensive. 
The basic way to calculate the probability of an expression is to apply the inclusion-exclusion rule repetitively which creates an exponential number of sub-expressions. There are however more efficient algorithms, such as the Heidtman~\cite{HEI89} algorithm or algorithms which make use of binary decision diagrams (BDD) \cite{BRY86,RAU03}, but the problem remains NP-complete.

\section{Approximating PAS on Complex Networks}
\label{sec:ApproximatingPAS}

We have shown that the exact degree of support calculations for PAS have non-polynomial complexity. 
Considering that the number of nodes affecting a node's rank can be as large as all the nodes in a complex network,  for many networks it is practically impossible to calculate the exact $dsp_i$ values. 

One possible way to control the complexity is to limit how far one goes back in the network for collecting support. 
We will use the term \textit{maximum order} of a supporting argument to refer to the number of link assumptions in the argument, as introduced in \cite{PIC98}. 
For example, in Example \ref{ex:SP} $SP(v_1)$ contains 
one supporting argument with 2 link assumptions ($a_2 \wedge l_{23} \wedge l_{31}$) and 
two others with only 1 link assumption ($a_2 \wedge l_{21}$ and $a_3 \wedge l_{31}$). 
Therefore the maximum order is 2.

Even calculations with a maximum order of 2 can be very difficult. 
Consider a citation network, for a paper we would have to consider the immediate citations, and then the citations to the citers. 
A paper can get more than 1000 citations and the citing papers may have citations to them. 
This would correspond to including the contributions of thousands of different papers in a $dsp$ calculation. 
We have used a BDD based implementation \cite{CET05} for exact $dsp$ calculations and 
we found that this calculation is impossible within realistic time/space limits. 
In \cite{PIC98} this is also reported as a problem where the author suggests use of a maximum order of 1 (using only immediate citers) where a higher order is not possible.

Although highly optimized algorithms in the future might get round to make such a calculation it is certainly not an easy task. Secondly, such a calculation with a maximum order 2 would fail to capture a more global picture in the network. Recall that one of the motivations behind the introduction of PageRank \cite{PAG98} was this.
 

For having a realistic chance to be applicable to ranking in very large complex networks an algorithm needs to have linear or close to linear time complexity and ideally utilize only local information to a node. In this section we will formulate such an algorithm. The ranking process will be viewed as a propagation of node probabilities over links in an iterative algorithm. There are two main challenges to consider, namely overestimation and cycles.

\subsection*{Overestimation}
We can make an exact calculation using only local information for a node if the supports of the citer nodes are disjoint. If we assume them to be disjoint when they are not, then we would overestimate the degree of support.
Let us detail this with an example. Consider Fig.\ref{fig:ExampleSimplePASgraph1}, the neighbors of node 1 are nodes 2 and 3. We know from Eq.\ref{eq:general_sp} the support for $v_1$ is: 
\begin{equation}
	SP(v_1) = a_1  \vee (SP(v_2) \wedge l_{21}) \vee (SP(v_3) \wedge l_{31})	
	\label{eq:sp_v1}
\end{equation}
If we assume $SP(v_2)$ and $SP(v_3)$ to be disjoint then we get $dsp_1'$ as below:
\[
dsp_1' = 1 - (1 - p(a_1 ))(1 - dsp_2 \; p(l_{21}))(1 - dsp_3 \; p(l_{31}))
\]
where we use inclusion-exclusion rule as in Eq.\ref{eq:sp_v1}. 
Using the values from our example we see that $dsp_1' = 0.5255$ compared to $dsp_1 = 0.5047$. 
Note the values are rather similar, and the difference is made by the overestimating of the effect of node 2.

This leads us to formulating the \textbf{common conjunction model} which uses a damping function $d_{c}(v_{i})$ to discount the possible effects of overestimation:
\begin{eqnarray*}
\label{eq:commonconjunction}
dsp_{i}' & = & 1-\left({1-p(a_{i})}\right) \\ 
         &   & \cdot \left( 1 - d_{c}(v_{i}) \left(1-\prod\limits _{j\in P_{i}}{(1-dsp_{j} \; p(l_{ji}))} \right)\right)
\end{eqnarray*}

\begin{figure}[hbt]
\center
\includegraphics[scale=0.5,keepaspectratio=true]{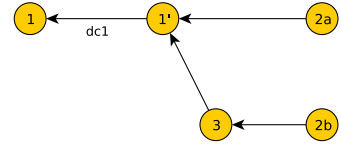}
\caption{\label{fig:ExampleSimplePASgraph1_trf_node1} Transformed graph as seen by node 1.}
\end{figure}

This is equivalent to doing a partial transformation on the immediate neighbors of a node, and accounting for the previous ``entanglement'' using an extra ``damping'' node, see Fig.\ref{fig:ExampleSimplePASgraph1_trf_node1} for a demonstration of the idea.

Recall that for small-world networks \cite{NEW03} it is shown that 
if vertex $i$ is connected to vertex $j$ and vertex $k$, 
then it is highly probable that vertices $j$ and $k$ are also connected.
Damping function is therefore used to counter the effect of the clustering.

We now formulate our first approximation method we name \textbf{ERank-0} as below:
\begin{eqnarray*}
\label{eq:erank0}
\widehat{dsp}_{i}^{k+1} & = & 1-\left({1-p(a_{i})}\right) \\
                        &   & \cdot \left(1-d_{c}(v_{i})\left(1-\prod\limits _{j\in P_{i}}
{(1-\widehat{dsp}_{j}^{k} \; p(l_{ji}))}\right)\right)
\end{eqnarray*}
where $ \widehat{dsp}_{i}^{k}$ is the $dsp$ estimate for node $i$ at iteration $k$ 
with the initial condition $\widehat{dsp}_{i}^{0} = 0$.
$ERank0(i)=\widehat{dsp}_{i}^{k}$ for a chosen number of iterations $k$. 
We can think of this as a series of approximations based on how far we go back in the network to look for support. 

ERank-0 produces gradually better estimates after each iteration. We typically use $d_{c}(v_{i}) = d_0$ where $d_0$ is chosen to minimize an objective function for a sample set of nodes in the network.

For Fig.\ref{fig:ExampleSimplePASgraph1} we see for example that using $d_0=0.95$ after three iterations $ERank0(v_1)=0.5127$, which is higher than the exact value but lower than what would be the if $SP(v_2)$ and $SP(v_3)$ were disjunct. We explore the effects of the damping values later on this section.

\begin{figure}[hbt]
\center
\includegraphics[scale=0.5,keepaspectratio=true]{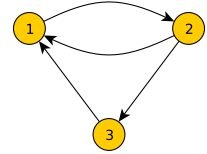}
\caption{\label{fig:ExampleGraphOfCycle} A simple network with a cycle.}
\end{figure}

\begin{figure}[hbt]
\center
\includegraphics[width=\columnwidth,height=\columnwidth,keepaspectratio=true]{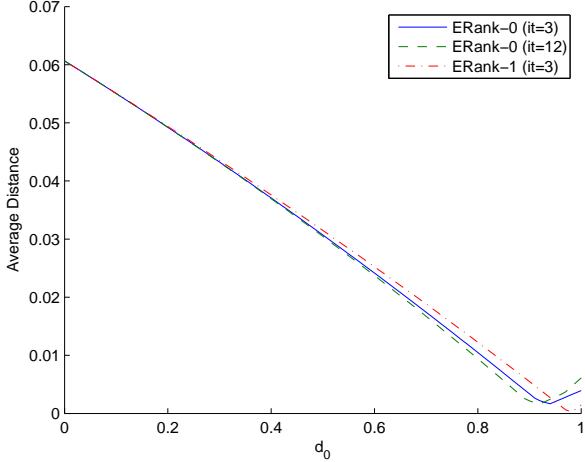}
\caption{\label{fig:Example2-distance} Average distance for various algorithms on Fig.\ref{fig:ExampleGraphOfCycle}.}
\end{figure}

\subsection*{Dealing with cycles}

ERank-0 is prone to deterioration of ranks in the presence of cycles between nodes. This effect is stronger with immediate cycles but still present when indirect cycles are present.

We formulate higher-order algorithms which avoid feedback for a given maximum number of links between nodes. Based on how many links they avoid the feedback, they are named; ERank-1 (avoids feedback between immediate neighbors, i.e. one link), ERank-2 (avoids feedback between nodes separated by another node, i.e. two links), or arbitrarily higher. ERank-0 has no such avoidance hence the ``0'' in the name. We also use \textit{ERank-N} to refer to avoidance of feedback from any possible length of links.
These higher-order algorithms (ERank-1 and above) use a message-passing scheme to avoid feedback from cycles by keeping a set of nodes which have already contributed to a calculation. Further details regarding ERank-N can be found in \cite{CET_ESP} and \cite{CET05}. Also, in \cite{CET05} we offer a formal treatment of the theoretical framework presented here, introducing the \textit{Entity Transitive Relation Implication} (ETRI) model for the mapping of a network into a PAS instance. In this previous work we present ERank as a special case tailored for the network ranking application of a general case algorithm named \textit{ETRI Support Propagation} (ESP). However we chose to use ERank throughout this article for the sake of simplicity also omitting other details that are not crucial.

For example in Fig.\ref{fig:ExampleGraphOfCycle} nodes 1 and 2 have an immediate cycle between them. Fig.\ref{fig:Example2-distance} shows how ERank-0 and ERank-1 perform when run on the network of Fig.\ref{fig:ExampleGraphOfCycle}. It plots the average distance for a given iteration:

\begin{equation}
d^k = 1/n \sum_{i=1}^{n} \left|\widehat{dsp}_i^k - dsp_i \right|
\label{eq:avg_dsp_distance}
\end{equation}

In this figure, we plot the results when ERank-0 is run for 3 iterations, and when it is run for 12 iterations. For comparison we also plot the results from ERank-1 at 3 iterations.

We observe ERank-0 algorithms with different iterations do comparably well, while ERank-1 outperforms others when $d_0$ is chosen correctly.

In our experimentation with the Reuters network we have not seen any significant improvements in estimation performances or ranking performances (as we introduce later) using these ``higher'' algorithms. This is probably because the Reuters network is undirected although we have not confirmed this. So we will not deal with the other ERank algorithms any further in this article due to space considerations.

\subsection*{Assigning node and link assumption probabilities}
For applying ERank algorithms in particular, and PAS based ranking/analysis in general one needs to assign prior probabilities to assumptions. We will deal with the two different types of assumptions in the network mapped PAS knowledgebase; node and link assumptions.

For the network of infection, the probability of the node assumption corresponds to the prior probability that an individual is infected. The probability of transmitting the infection is represented by the link assumption probabilities.

If such prior probabilities for a relation in the network are known they may be useful. 
Lack of such data does not make the analysis impossible though. 
In this work we will use $p(a_i)=1/n$ where $n$ is the number of nodes. 
In the evidence theory (DST) interpretation, 
this corresponds to assuming that at least one node in the network has the analyzed property. 
It can be thought of as a \emph{minimal evidence} or the most conservative assumption to make about the network before analyzing it for a property.

If prior link probabilities are not known, we can not offer a similarly simple assignment for link probabilities. 
Instead a range of values, such as conservative estimates depending on the relation can be used as we will show below. 
We use $p(l_{ij})=p_{l0}$ for all $i, j$ where $ p_{l0}$ is a model parameter and various values of it are investigated.

When applying ERank algorithms on the Reuters network we will use the transitive relation: ``(if person A links to B) person B is influenced by person A''. So, we will interpret our results to yield the posterior probability of a person being influential. 

\subsection*{ERank algorithms for approximating $dsp$ values}

For successfully applying ERank algorithms, one needs to choose the number of iterations to run and what damping function or constant to use. 

Let us use $\iota$ to denote the number of iterations. For ERank-0 for a given $\iota$ the corresponding maximum order approximated is $\iota - 1$. It is not hard to see how this is. Each iteration after the first one generates approximations for an additional order of support compared to the previous iteration. Therefore the highest number of potentially useful iterations is limited with the diameter of the network. Using additional iterations do not necessarily create better approximations though and it depends on the structure of the network what value number of iterations is the most suitable. A way to decide on an $\iota$ is to take into account what the maximum contribution a supporting argument of the corresponding order would be, and if there are significantly many supporting arguments to make a difference. For example, when the algorithm is run for 6 iterations than the maximum order of corresponding supporting arguments is 5. Assuming $p_{l0}=0.2$ gives $0.2^5=3.2\cdot 10^{-4}$ as the maximum contribution a supporting argument of order 5 would give, compared to $0.2$ for immediate neighbors of a node. Note also that it is known in the small-world network model the average of the distances between nodes is unusually low compared to a random network \cite{NEW03}. This can serve to limit the maximum number of iterations needed even for a very large network.

In this work we use a constant damping function $d_0$ although it is possible to come up with a different heuristic function. The choice of the damping constant relies similarly on the structure of the network. In this section we will use Eq.\ref{eq:avg_dsp_distance} as an objective function and plot different approximation results using it.

As we have argued earlier, the exact $dsp$ value of a node may be prohibitively hard to compute. On the Reuters network we have been able to compute the exact $dsp$ values of nodes up to different maximum orders ranging from one (just the immediate neighbors) to 11. We use as many as possible of these as sample sets to plot the average distance using Eq.\ref{eq:avg_dsp_distance}. For example when comparing against ERank-0 run with 6 iterations, we use all of the sample set for which we could calculate the $dsp$ values using the corresponding maximum order of 5. We do not include nodes without any links in these calculations.

\begin{figure}[hbt]
\center
\includegraphics[width=\columnwidth,height=\columnwidth,keepaspectratio=true]{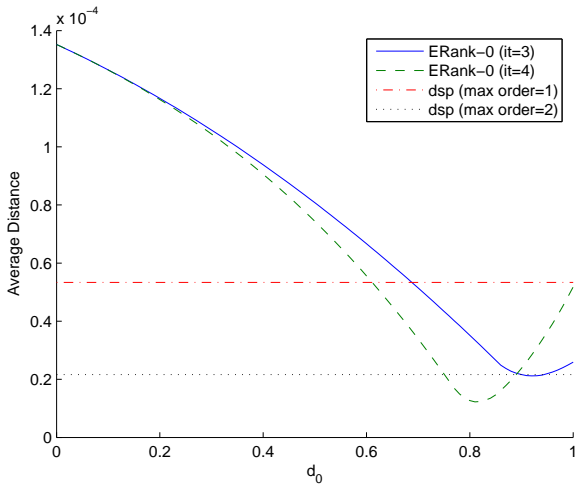}
\caption{\label{fig:ERank0_vs_dsp} ERank-0 and $dsp$ computations approximating $dsp$ computations of maximum order 3 which is represented by $y = 0$.}
\end{figure}

In Fig.\ref{fig:ERank0_vs_dsp} we consider the average distance on the Reuters network where comparisons are made against $dsp$ calculations with a maximum order of 3. It contains the plots of ERank-0 for $p_{l0}=0.2$ and $p(a_i)=1/n$ using 3 and 4 iterations for the damping constant range $[0,1]$ along with corresponding $dsp$ computations using maximum orders of 1 and 2. The results are offset in reference to $dsp$ with maximum order 3 which is represented by the line $y = 0$. We observe that when ERank-0 has a good damping constant it can outperform exact $dsp$ calculations of maximum order 2.

\begin{figure}[hbt]
\center
\includegraphics[width=\columnwidth,height=\columnwidth,keepaspectratio=true]{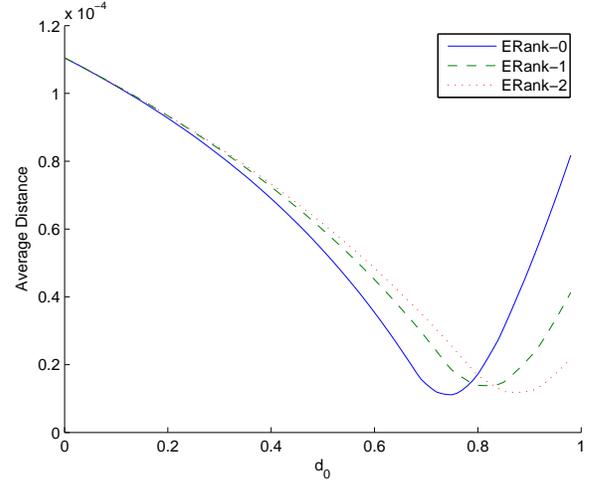}
\caption{\label{fig:eranks-it6} Comparison of different ERank algorithms corresponding to a maximum order of 5.}
\end{figure}

Similarly, in Fig.~\ref{fig:eranks-it6} we use the same probability values as in Fig.\ref{fig:ERank0_vs_dsp} to compare how different ERank's perform on the Reuters network. Using Eq.\ref{eq:avg_dsp_distance} we plot ERank results comparing them to $dsp$ computations with a maximum order of 5. ERank-0 appears here to perform as good as the higher order ERank algorithms. As we have argued above we believe this is because the conversion from undirected to directed network places cycles for all the nodes although we have not validated this yet. 

Finally, observe that when computing ranks for ERank-0 one calculation is made over every link per iteration. So ERank-0 has a linear time complexity $O(l)$ with the number of links $l$ per iteration.

\section{A Performance Evaluation of Ranking Algorithms}
\label{sec:ReutersPerformanceEval}
In this section we will propose a method to compare the performances of different ranking algorithms on the Reuters network and then present a study of the performances of a number of well-known algorithms comparing them to ERank algorithms. 

\subsection{Assessing importance of nodes}
We will link the importance of a person in 1987 to importance today. We will see how well a person in the Reuters collection is represented in today's English Wikipedia and compare that with the rankings. Part of this study appeared before in~\cite{OZG08}. 

For assessing the validity of our results we have used a crawler to look up if a given person has an English Wikipedia page~\cite{Wikipedia}. 
We have interpreted this as an indication that a given person is important today in a general global sense. 
This would have an English speaking world bias and may not necessarily be a truly objective measure. 
However Reuters also being an English source and English being the closest there is to a truly global language, this measure should function at least to a reasonable extent. 
Our basic assertion here is that if a person was important back in 1987 when the Reuters articles were being published, then s/he would still be important today. 
The 20 years passed since then can make a ``time's judgment'' on who were truly important at the time. It is possible however other people in those articles unimportant or unforeseeable at the time will have gained importance. Similarly some who were not very important from a Reuters reporting perspective can actually be important individuals for different reasons.  
Combined, these would mean that the assessment power of the algorithms would be limited in discovering all those who are important, however this analysis should be reasonably good enough to penalize ``false positives'' which the algorithms would mark as important but were really not as such. 

Using the crawler results we have constructed the function: \textit{``has a page''} $H(i)$ which is 1 if there is any Wikipedia page for a given person $i$, 0 otherwise. Of the 5,249 persons in the network we find that 1,440 have a Wikipedia page. In the rest of this section we will use this function as apriori information on the importance of nodes and perform a comparative study of the algorithms. 
Table~\ref{cap:top20acount} shows the top 20 people when ranked according to article count values. 
Having a glance at this table can serve as a basic reality check for the utility of our defined functions. 
For example we see that most of the people we could expect to have high importance have $H(i)=1$; President of USA, Prime Minister of Japan, Secretary of State of USA. 

\begin{table}[hbt]
\caption{Top-20 persons in article count\label{cap:top20acount}.}
\center
\begin{tabular}{|c|c|c|c|}
\hline 
	person & a. count & $H(i)$ & notes  \tabularnewline
\hline 
\hline 
	r.reagan & 493 & 1  & President\tabularnewline
	j.baker & 212 & 1  & Treasury Secretary\tabularnewline
	y.nakasone & 112 & 1  & Prime Minister, Japan \tabularnewline
	p.volcker & 109 & 1  & Ch. Fed. Resv. Board\tabularnewline
	k.miyazawa & 86 & 1  & Finance Minister, Japan \tabularnewline
	c.yeutter & 85 & 1  & Trade Representative \tabularnewline
	n.lawson & 66 & 1  & Chan. Exchequer, UK \tabularnewline
	d.funaro & 58 & 0  & Fin. Minister, Brazil \tabularnewline
	r.lyng & 57 & 1  & Agriculture Secretary \tabularnewline
	g.stoltenberg & 55 & 1  & Fin. Minister, W.Germ. \tabularnewline
	g.shultz & 50 & 1  & Secretary of State \tabularnewline
	m.thatcher & 50 & 1  & Prime Minister, UK \tabularnewline
	e.balladur & 48 & 1  & Fin. Minister, France \tabularnewline
	j.wright & 47 & 1  & W.H. Speaker, Texas\tabularnewline
	s.sumita & 44 & 0  & Bank of Japan Gov.\tabularnewline
	m.baldrige & 42 & 1  & Commerce Secretary\tabularnewline
	m.fitzwater & 40 & 1  & W.H. Speaker\tabularnewline
	a.greenspan & 39 & 1  & Ch. Fed. Resv. Board \tabularnewline
	j.ongpin & 36 & 0  & Fin. Secr., Philippines \tabularnewline
	j.sarney & 36 & 1  & President, Brazil \tabularnewline
\hline 
\end{tabular}
\end{table}

\subsection*{Performance as clustering validity}
The function $H(i)$ can be thought as placing each node in one of the two classes $0$ and $1$, i.e. those with and without English Wikipedia pages. Hence this becomes a clustering problem with an external criteria. We would ideally like an algorithm to rank all the persons labeled as $H(i)=1$ higher than the ones labeled with $0$, thus giving us a perfect separation of the collection into two clusters. There is a well-known statistic named ``Hubert's gamma'' which is used for assessing cluster validity in this class of problems~\cite{JAIN88}. Mathematically stated Hubert's gamma is:
\begin{equation}
\Gamma  = \sum\limits_{i = 1}^{n - 1} {\sum\limits_{j = i + 1}^n {X\left( {i,j} \right)Y\left( {i,j} \right)} } 
\end{equation}
where 
\begin{equation}
Y(i,j) = 
\left\{ 
\begin{array}{rcl} 
0 & \mbox{   } & \mbox{if $H(i)=H(j)$}  
\\ 1 & \mbox{   } & \mbox{otherwise}
\end{array}
\right. 
\end{equation}
and $X(i,j)$ is the distance between the two nodes. $X(i,j)$ is usually the Euclidian distance on the ranks. Let us use $\rho(i)$ to denote the rank value given to node $i$ by the ranking algorithm $\rho$. Then the Euclidian distance function is: $X(i,j)=\left|{\rho(j)-\rho(i)}\right|$. The $\Gamma$ statistic measures the degree of linear correspondence between the entries of $X$ and $Y$.

The power of a statistical test is in establishing how unusual a given ordering is. To do this we come up with a null hypothesis $H_0$ which is a statement of ``no structure''. The $H_0$ for $\Gamma$ is called the ``random label hypothesis''(RLH) which postulates that all permutations of the labels on $n$ objects are equally likely. We establish a distribution for $H_0$ using Monte Carlo sampling creating random permutations of node labels on our collection (we shuffle the node labels and calculate corresponding $\Gamma$s). For $\Gamma$, the higher the value the more likely that a given labeling is unusual. We use the RLH distribution to compare with the $\Gamma$s obtained from our algorithms, and if we find these $\Gamma$s to be unusually large then we can conclude the algorithm is successful.

Since we wish also to compare the performances of the different algorithms, we have used the positions assigned by the algorithms to a node instead of the rank values. This way we make the $\Gamma$ values obtained directly comparable. For example $X(i,j)$ would be defined as 
\[
X(i,j)=\left|{Pos_{\rho}(j)-Pos_{\rho}(i)}\right| 
\]
where $Pos_{\rho}(i)$ is the position given by the algorithm to node $i$ according to $\rho$. This however brings another problem when ranking algorithms assign the same rank value to a large set of nodes: two nodes with the same rank can have positions which are far apart thus being ranked very differently in terms of positions despite being equivalent in actual ranks. To overcome this problem we did a random sampling of different orderings in which nodes with equal rank values are shuffled into random positions between each other for each calculation of $\Gamma$. This for example then gives our distance function $X(i,j)$ for $\rho$ as:
\begin{equation}
X(i,j)=\left|{\overline{Pos_{\rho}(j)}-\overline{Pos_{\rho}(i)}}\right|
\end{equation}
where $\overline{Pos_{\rho}(i)}$ is the average value of $Pos_{\rho}(i)$ obtained after the random sampling.

Hubert's $\Gamma$ combined with the $H(i)$ thus gives us a statistical test to compare the performances of any ranking algorithm on the Reuters network.

\subsection*{Performance results}
We have run the ERank algorithms ERank-0, ERank-1 and ERank-2 on the Reuters network. We use the results from following algorithms to compare:
\begin{itemize}
	\item \textit{Article count}, is the number of articles a person appears in.
	\item \textit{Degree} is the number of people a person got associated with in the collection, i.e. the link count on the node (in the undirected network).
	\item \textit{Closeness}, calculated using the undirected unweighted network.
	\item \textit{Betweennes}, calculated using the undirected unweighted network.
	\item \textit{PageRank}, is the PageRank of a node using $d = 0.5$. For application we have converted the undirected network to directed by replacing each edge with arcs in both directions.
\end{itemize}

\begin{figure}[hbt]
\center
\includegraphics[width=\columnwidth,height=\columnwidth,keepaspectratio=true]{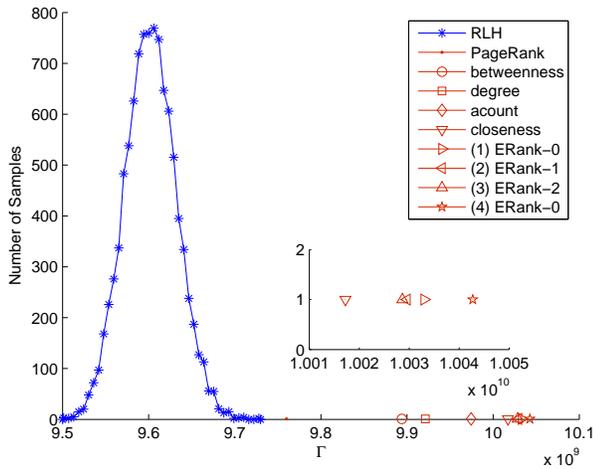}
\caption{\label{fig:gammas_RLH_and_all_algs} 
$\Gamma$s for algorithms and the RLH. $1.001-1.005 \times 10^{10}$ region  is expanded in the inset.}
\end{figure}

The $\Gamma$s for all the algorithms are on Table \ref{cap:gammas}, these and later results on the figures are obtained averaging the calculations of 100 samples. 
Fig.~\ref{fig:gammas_RLH_and_all_algs} gives how the $\Gamma$ values for the RLH and the algorithms relate. For this experiment we have used 10000 samples for calculating the RLH distribution assigning them to 40 bins in an histogram.

\begin{table}[hbt]
\caption{$\Gamma$s for different algorithms\label{cap:gammas}.}
\center
\begin{tabular}{|c|c|c|}
\hline 
	algorithm & $\Gamma$ & parameters \tabularnewline
\hline 
a. count   &$9.974 \cdot 10^{09}$& \tabularnewline
degree     &$9.921 \cdot 10^{09}$& \tabularnewline
betweenness&$9.894 \cdot 10^{09}$& \tabularnewline
closeness  &$1.002 \cdot 10^{10}$& \tabularnewline
PageRank   &$9.760 \cdot 10^{09}$&$d=0.5$ \tabularnewline
(1) ERank-0&$1.003 \cdot 10^{10}$&$\iota=6$, $p_{l0}=0.2$, $d_0=0.7$\tabularnewline
(2) ERank-1&$1.003 \cdot 10^{10}$&$\iota=3$, $p_{l0}=0.2$, $d_0=0.8$\tabularnewline
(3) ERank-2&$1.003 \cdot 10^{10}$&$\iota=2$, $p_{l0}=0.2$, $d_0=0.9$\tabularnewline
(4) ERank-0&$1.004 \cdot 10^{10}$&$\iota=12$, $p_{l0}=0.1$, $d_0=0.3$\tabularnewline
mean RLH   &$9.599 \cdot 10^{09}$& \tabularnewline
\hline 
\end{tabular}
\end{table}

We find that all the algorithms in fact give a valid clustering as the $\Gamma$s produced by the algorithms are higher than the whole sampling collection for the RLH. For Monte Carlo sampling, when $m$ is the sample size, and if $\Gamma_0$ is among the $k$ largest of the $m$ values in the sample set, then the probability of incorrectly rejecting $H_0$ when it is true is $\alpha = k/m$. $k$ is usually chosen higher than 5 \cite{JAIN88}, so for this experiment using $m = 10000$ and $k = 10$ we get the level of significance as $ \alpha = 0.001$ which is a high confidence level.

It is not a surprise that all the algorithms yield a valid clustering given that these are widely used in different applications. However we can distinguish between the comparative performances of the algorithms statistically, as to how unusually good their given results are. We observe that when accordingly parameterized ERank outperforms all other algorithms.

\section{Choosing ERank parameters}
\label{sec:ChoosingERankParams}

A successful application of ERank depends on choosing various parameters. Firstly, for constructing the PAS instance, one has to choose the probabilities of assumptions; $p(a_i)$ and $p(l_{ij})$ based on the transitive relation used. Then, a damping function (e.g. the constant damping function $d_0$) and the number of iterations $\iota$ has to be chosen. All of these have complex interactions and it is not always clear how they relate to each other and the algorithm performance in general. In this article, we have employed a constant node assumption probability function $p(a_i)=1/n$ and a link assumption probability function $p(l_{ij})=p_{l0}$, along with the constant damping function $d_0$. In this section we will briefly explore how these different parameters interact and affect the algorithm performance as indicated by $\Gamma$ in the Reuters network.

\begin{figure}[hbt]
\center
\includegraphics[width=\columnwidth,height=\columnwidth,keepaspectratio=true]{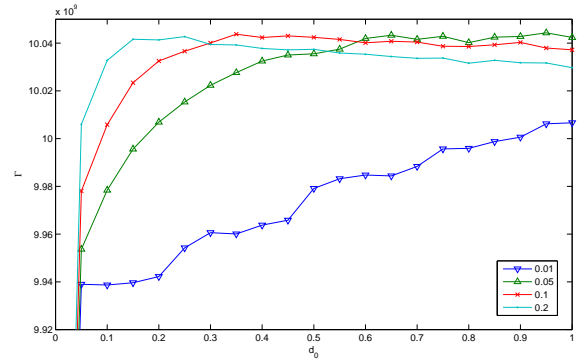}
\caption{\label{fig:ERank0-different-pl0} $\Gamma$s using ERank-0 for different $p_{l0}$ and $d_0$ values using $\iota=6$.}
\end{figure}

\begin{figure}[hbt]
\center
\includegraphics[width=\columnwidth,height=\columnwidth,keepaspectratio=true]{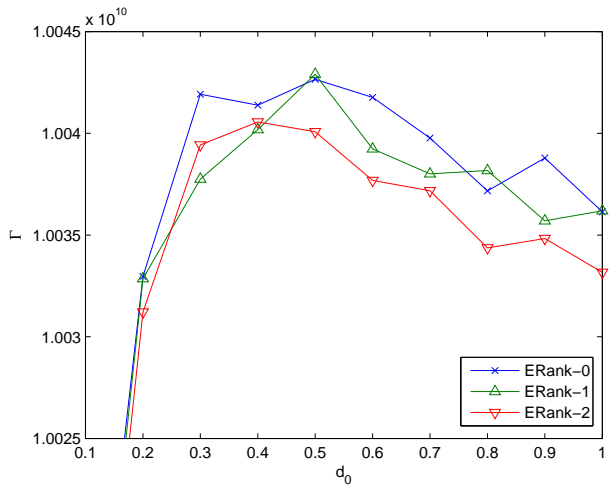}
\caption{\label{fig:DifferentERanks} $\Gamma$s using different ERank-N algorithms for $p_{l0}=0.1$ and $0.1 \le d_0 \le 1.0$ corresponding to a maximum order of 5.}
\end{figure}

\begin{figure}[hbt]
\center
\includegraphics[width=\columnwidth,height=\columnwidth,keepaspectratio=true]{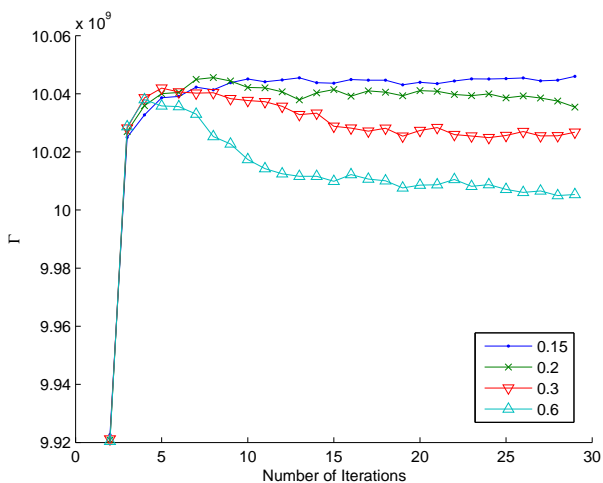}
\caption{\label{fig:ERank0-Iterations} $\Gamma$s using ERank-0 for $p_{l0}=0.1$ with increasing number of iterations for different $d_0$ values.}
\end{figure}

In Fig.~\ref{fig:ERank0-different-pl0} we see how different $p_{l0}$ values affect $\Gamma$ values for different $d_0$ values using ERank-0. As can be seen, some $p_{l0}$ values result in a wider range of $d_0$ values where high $\Gamma$s are obtained. The optimal $d_0$ values are much lower for the $\Gamma$ calculation as compared to what is discovered in the approximation section (e.g. for $p_{l0}=0.2$). This may be a shift due to the change in the objective function and the use of positions and not actual values. Also the nodes in the dense areas of the network may shift the average clustering to a higher degree. Another observation is how the results are robust for a range of $d_0$ and $p_{l0}$ choices. Fig.~\ref{fig:DifferentERanks} shows how different ERank algorithms yield results. In line with the approximation results, ERank-0 is the best performer by a small margin. Finally Fig.~\ref{fig:ERank0-Iterations} plots how $\Gamma$s change by increasing iterations for different $d_0$ values. Usually the $\Gamma$ values start dropping around iteration 4-8, however an interesting observation here is that $d_0=0.15$ appears unnaturally stable. This may be because of $d_0$  compensating also for immediate cycle effects.

\section{Conclusion}

We have introduced a family of novel rapid approximation algorithms for applying a PAS based modeling and ranking to large complex networks (particularly small-world model networks). As far as we are aware, it is the first of its kind that is both practically applicable to large networks and formally founded in a quantitative reasoning framework. A problem known to be NP-complete is approximated using linear and near linear time algorithms for this specialized application domain. Thus ERank enables the use a new paradigm in addition to the Markov (random surfer) model for ranking probabilistically.

We have explored various issues for a sound application of the algorithm on the Reuters~\cite{OZG08} network. These include; the choice of a damping function, assigning the prior node and link assumption probabilities and choosing the number of iterations. 

We propose a statistical test to compare the performances of any ranking algorithm on the Reuters network using a clustering validity test. We apply a number of well-known algorithms and compare their results with ERank algorithms. When ERank algorithms are parameterized accordingly, they perform better than the other algorithms. An unexpected finding was that PageRank was the worst performing of the algorithms considered (more on this in \cite{OZG08}). This may be related to the conversion from the undirected network to directed.

Our experimentation reports good performance for a wide range of parameters. This is good in the sense that ERank appears to be robust. Also, it is possible to interpret this as the test not being able to distinguish performance results above a certain precision or threshold, although it was good enough to uncover performance differences between the various algorithms.

The superior performance of ERank may be attributed to a global character present in the final ranks. For example in a given network, a node in a ``dense'' area will surely be ranked highly despite possibly very intricate details of linking between the nodes. Once the obvious source of distortions are removed (e.g. immediate cycles) and an expected clustering is accounted for (i.e. the damping function) the ``big picture'' can be obtained correctly despite many possible distortions.

ERank as we apply it, is susceptible to various sorts of manipulations as a ranking algorithm. For example it would not be able to discover an unusual overestimation caused by a high rank source behind a facade of immediate neighbors. This is by design, that we have used a constant damping function. One may need to come up with a better heuristic function or a combination of exact and approximate algorithms can be used. On the other hand, it is a global ranking algorithm like PageRank and would have resistance to manipulation in this sense. Therefore testing its robustness against manipulation is a possible future research direction.

A problem with this experimentation is the conversion from undirected to a directed graph. While interesting as an experimentation on an (essentially) undirected graph, using the Reuters network we were not able to test our algorithms on a truly directed network. It remains as future work to apply ERank on a truly directed graph and evaluate performance against apriori information. On such a graph we would expect ERank-N with $N > 0$ to outperform ERank-0.

Also as future work, it would enhance the reliability of the prior information to include information from Wikipedias of different languages, as well as using other references sources.

What we present here attempts to nominate ERank as a good algorithm for at least some ranking applications. Possibly much more needs to be done to establish how different ranking algorithms including ERank compare with each other for different applications. In this regard, given ERank's theoretical soundness and the superior performance in this experimentation, we hope to stimulate further research and interest in this direction.



%

\bibliographystyle{IEEEtran}
\bibliography{ERank-Reuters}

%







\end{document}